\documentclass[12pt, a4paper, notitlepage, oneside]{article}
\usepackage{graphicx}
\usepackage[utf8]{inputenc}
\usepackage{amsmath}
\usepackage{amssymb}
\usepackage{hyperref}
\usepackage{algorithm}
\usepackage{algpseudocode}

\title{Fair Adversarial Networks}

\author{George \v{C}evora   \\ illumr Ltd., London, United Kingdom \\ \href{mailto:george.cevora@illumr.com}{george.cevora@illumr.com}}  

\date{}

\begin{document}

\maketitle

\abstract{ The influence of human judgement is ubiquitous in datasets used across the analytics industry, yet humans are known to be sub-optimal decision makers prone to various biases. Analysing biased datasets then leads to biased outcomes of the analysis. Bias by protected characteristics (e.g. race) is of particular interest as it may not only make the output of analytical process sub-optimal, but also illegal. Countering the bias by constraining the analytical outcomes to be fair is problematic because A) fairness lacks a universally accepted definition, while at the same time some definitions are mutually exclusive, and B) the use of optimisation constraints ensuring fairness is incompatible with most analytical pipelines. Both problems are solved by methods which remove bias from the data and returning an altered dataset. This approach aims to not only remove the actual bias variable (e.g. race), but also alter all proxy variables (e.g. postcode) so the bias variable is not detectable from the rest of the data. The advantage of using this approach is that the definition of fairness as a lack of detectable bias in the data (as opposed to the output of analysis) is universal and therefore solves problem (A). Furthermore, as the data is altered to remove bias the problem (B) disappears because the analytical pipelines can remain unchanged. This approach has been adopted by several technical solutions. None of them, however, seems to be satisfactory in terms of ability to remove multivariate, non-linear and non-binary biases. Therefore, in this paper I propose the concept of \emph{Fair Adversarial Networks} as an easy-to-implement general method for removing bias from data. This paper demonstrates that Fair Adversarial Networks achieve this aim.
}

\newpage
Unfair treatment of individuals enacted by automated decision-makers has recently attracted a large amount of attention \cite{o2016weapons,eubanks2018automating,zou2018ai}. This problem is, however, not limited to automated decision-making; all Data Analytics [DA] and Machine Learning [ML] has the potential to lead to biased conclusions and therefore enact discrimination. The prominence of automated decision-makers in the discussion of discrimination is mostly due to two factors: 1) machine learning models are hard to scrutinise and therefore lack the human oversight that otherwise could prevent discrimination, and 2) automated decision-makers are easy to test and therefore they can be relatively easily proven to be biased. Many conventional uses of Data Analytics can lead to same problems though - for instance the analyst may conclude that individuals from particular postcodes are less credit-worthy. If that judgement is based on biased data and those postcodes are predominantly home to individuals of a specific race, this amounts to racial discrimination despite race not being considered by the analyst at all. The methods presented in this paper aim to prevent these exact situations and at the same time provide a generic easy-to-use solution accessible to any data analyst.

Discrimination is a result of past biased human judgements that make-up the datasets,  or by biased human judgements which shape the composition of the dataset \cite{barocas2016big,caliskan2017semantics}.
This paper focuses on illegal discrimination; however, the findings are applicable to all types of bias, including biases that would not be generally seen as unfair. For instance in the case of dealing with HR data collected across a number of regional offices, the bias of the region might be undesirable in employee evaluation.

It has been reported \cite{turner1999mortgage} that it is harder for African-Americans to obtain loans than for their white-American counterparts with the same repayment ability.  
%This concerns loans approved by human decision-makers as well as those approved by an automated decision-maker. 
%The former is a simple case of racism, while the second is a consequence of data riddled by racism that was used for training of the Machine Learning [ML] models in question. 
As race is one of the protected characteristics under US law, use of such a system to perform decision-making is illegal. However, this problem does not only concern ethics or legality. In fact, there is a solid business case for fairness. Let us consider a case in which two individuals with equal repayment ability apply for a loan but one of them is rejected based on discrimination against a group they are a member of. Besides being unfair and illegal this also makes a lost profit for the lender in question.

There are two main scenarios in which racism creeps into DA. The more straightforward scenario occurs whenever the objective of the DA exercise is to predict (biased) human judgement. The need to remove the past human bias in this scenario is widely acknowledged, yet often ignored. Examples of this issue come from many of the companies that use DA or  ML for hiring. They may try to link a candidate's characteristics such as CV, performance on aptitude tests, or even appearance to figure out how well the candidate will perform in the job. However, they almost never link that information with actual performance of people hired in the past. They are more likely to simply check whether the individuals with particular characteristics were hired or not, which is the decision of a potentially biased hiring manager. This approach is pragmatic - evaluating performance is not easy and can be only done for individuals who were actually hired, not for all applicants, thus severely reducing the size of the dataset.

The fairer and more accurate option is to use the actual performance of the individuals hired on the job to guide future decision making. However, this data too can be biased. The reason for this is that the datasets used for this task are almost never a uniform sample from the population. For instance, women have lower chances to be hired for many roles, therefore they are less likely to be included in the dataset used to train the automated decision-maker. This results in a situation in which the dataset used is systematically different to the population, where the difference may increase over time - an ubiquitous sight in DA/ML. This approach too offers a biased view of the population.

%problem 2 - alternative solutions
Several attempts to address this issue have been made in the past \cite{feldman2015certifying, hardt2016equality,zemel2013learning}. However, a complete lack of consensus on which of these methods should be the gold standard  led to the current status quo in which data analysts or ML engineers decide themselves how to address this problem \cite{barocas2016big}. 

The bias-removal methods can be broadly divided into two categories: 1) methods seeking statistical parity between groups of individuals that differ on their protected characteristic; 2) methods that seek to remove information about the protected characteristics in the datasets used for ML.
	%2a - statistical parity

\subsection*{Statistical Parity}
Ensuring statistical parity between groups, such as equal access to opportunity, is one of the most popular bias-aware method in DA/ML \cite{kamishima2012fairness}. The fundamental problem with this approach is the necessity to explicitly define fairness. Unfortunately there is a complete lack of consensus on this issue. In fact Berk and colleagues \cite{berk2017fairness} identify at least six different notions of fairness in the context of penal systems. Many of these notions are not mutually compatible \cite{chouldechova2017fair}, therefore the bias-aware data analysts have to pick what fairness means to them. This principle is fundamentally flawed as the notion of fairness should not be subjective. Unfortunately, while the legal system requires  fairness in respect to protected characteristics (e.g. US - Human Rights Act, EU - European Convention on Human Rights), it doesn't provide sensible\footnote{The U.S. Supreme Court asserted the principle of disparate impact, while resisting any sort of mathematical definition (p. 5,  \cite{feldman2015certifying}). Other bodies such as the Equal Employment Opportunity Commission \cite{equal1990uniform} have provided some guidance, but these are not legally binding and leave a lot to be desired.} guidance on what type of statistical parity is required. 

% difficult optimisation
A further big hurdle for achieving fairness through statistical parity is the difficulty of implementing it. Whilst matching the means and standard deviations between the groups is easy\footnote{Just a simple elastic transformation of one group's outcomes to match the outcomes of the other group.} it is also a fairly weak notion of fairness \cite{chouldechova2017fair}. Most other notions of fairness require inclusion of a parity optimiser into the DA/ML model - adding a fairness term to the loss function. This is clearly outside the area of expertise of most data analysts, as they do not tend to write their own loss functions. It also renders most standard DA functions or packages, such as scikit learn, obsolete as there usually are no reasonable options to customize loss functions. Therefore this approach causes a significant disturbance to existing DA pipelines and significantly increases the time commitment from the analysts, as well as the technical prowess required for such a role.
	
	%problem 2 - accuracy vs ethics	
Furthermore, it should not be forgotten that often the data used for ML does not represent the population well. Optimising for parity on training data doesn't guarantee parity at the population level when the training sample is heavily biased.
		
\subsection*{Removing protected characteristics}
	%2b - removing the information
An approach that avoids many of the issues of the statistical parity is to remove the information about protected characteristics from the dataset. The logic behind this issue is beautifully elegant; an agent that is not aware of the protected characteristics can not discriminate based on those characteristics. Unfortunately, the realisation of this principle often fails to deliver on its promises.

%removing columns
An approach that is sometimes applied to remove information on protected characteristics is a simple exclusion of these labels from the dataset. While it is widely recognized \cite{o2016weapons} that this approach is highly flawed, it is still widespread across the industry (author's anecdotal evidence). The logic behind the failure is simple, the information about the protected characteristics is partially contained in the other characteristics relevant for the decision at hand \cite{pedreshi2008discrimination}. These cases include racially segregated neighbourhoods where the postcode reveals the race of an individual, or professions that have traditionally high gender imbalance which reveal gender of the individual. At the same time, home address or previous profession are highly desirable information for a whole range of data applications; therefore can not be excluded from the dataset. It is therefore apparent that this approach cannot lead to fair and accurate DA outcomes.

A more sophisticated application of the same principle is manipulating the dataset in a way that removes the ability to predict the protected characteristics for the individuals in the dataset from the rest of the data. This approach has been developed over time to a great success \cite{bolukbasi2016man,hajian2011discrimination,hardt2016equality,zemel2013learning}; however, some shortcomings are still apparent: 1) the relationship between protected and other characteristics is certainly not linear like most methods assume; 2) even in cases where a variable seems to not relate to the protected characteristic, it may do so via an interaction with another variable; 3) none of the methods can address continuous characteristics (e.g. age) and even multiple categories pose significant problems \cite{bolukbasi2016man}. Proposing a novel method that  addresses these issues is the main objective of this paper.

\section{Fair Adversarial Networks}
\label{sec:FANs}
A number of methods of removing bias from datasets have been devised, however they generally fall short at removing non-linear, non-binary and/or multivariate biases.
To address the problems of the current methods this paper introduces a novel concept of \emph{Fair Adversarial Networks} [FANs]. Like Generative Adversarial Networks [GANs] \cite{goodfellow2014generative} this method consists of two networks with different objectives trained iteratively. FANs are a system that creates an unbiased version of a dataset that can subsequently be used with any analytical tools. There are two main components to this system 1) an \emph{autoencoder} \cite{rifai2011contractive}  function $y=\rho(x,W_A)$ that provides $y$ a reconstruction of data $x$ given autoencoder weights $W_A$, and 2) a \emph{Racist Network} that provides estimate $\hat{r}$ of the true protected characteristics $\bar{r}$  (\emph{race} in this example) from $y$. 

\begin{figure}
	\centering
	\includegraphics[width=10cm]{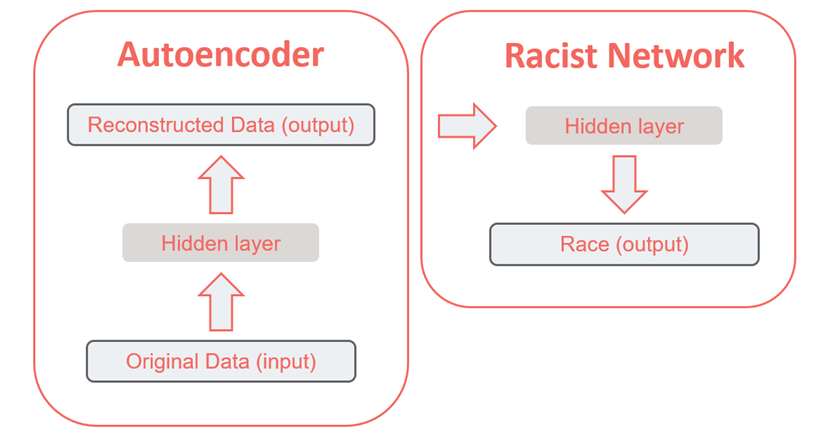}
	\caption{Architecture of a Fair Adversarial Network. Two networks are trained iteratively: the Racist network minimizes error on predicting race (or other protected characteristic of interest) while the autoencoder jointly minimizes its reconstruction error and the predictive capability of the Racist network.}
	\label{fig:architecture}
\end{figure}

The measure of bias we wish to minimize\footnote{Please note that the the cross entropy function $D$ is only appropriate to evaluate classifiers, not regressors, where a different objective function needs to be used.} is the true performance, such as the cross-entropy function $D$, of the Racist Network after full training $\bar{D}$ given the autoencoder weights at given epoch\footnote{$\bar{D}$ is identical to $D$ when optimal training schedule of the Racist Network has been concluded.}.
The main problem of this approach is the complexity of such a bias measure. The measure of bias is required at each epoch on which the autoencoder is trained, but it is not easy to obtain. Unlike the discriminator accuracy in GANs, the only way find the bias of encoded data is to fully train the racist network - to measure the true potential of the data to reveal the protected characteristic. Without full training, there is no guarantee that a failure of the racist network to predict the protected characteristic (detect bias) is not simply due to recent changes in the encoding that the racist network has not been able to adapt to yet. Therefore, to find the bias of the data exactly, we would have to perform lengthy full-training of the racist network on each epoch of the autoencoder training, clearly making the algorithm impractical. Furthermore, a guarantee of optimality for the network hyperparameters would be needed.  

We have developed a method for approximating the fully-trained performance from a single forward pass through the network $\hat{D}(\hat{r},\bar{r})$ which is the core of the autoencoder loss function specified by the equation \ref{eq:autoencoder}. Further developments to this measure that are required include a general mechanism to stabilise the adversarial training and a number of regularisers $\mathcal{R}$ that ensure quality of the final data encoding. While these developments cannot be shared publicly as they are core to the intellectual property of illumr Ltd., standard approaches are enough to replicate debiasing process on a one-off basis given sufficient hyperparameter tuning.

Formally, autoencoder minimizes loss function 
\begin{equation}
\mathcal{L}_A(x,r,W_A,W_R) = MSE(y,x) + c\hat{D}(\hat{r},\bar{r}) + \mathcal{R}
\label{eq:autoencoder}
\end{equation}
where $MSE(y,x)$ is the reconstruction error (Mean Square Error) of the autoencoder, $W_R$ the Racist Network weights, and $c$ is a constant balancing the individual terms of loss function. The Racist Network optimizes the appropriate loss function such as
\begin{equation}
 \mathcal{L}_R(y,\bar{r},W_R) = D(\hat{r},\bar{r}),
\label{eq:racist}
\end{equation}
which is a cross entropy function of two vectors $\bar{r}$ and $\hat{r}$.
As mentioned before this cost function ($D$) is different to the estimate of the performance of the Racist Network as minimized by the autoencoder ($\hat{D}$). This is because $D$ is a bad approximation of performance after full training $\bar{D}$.

This system of two networks is trained in an iterative adversarial fashion similar to GANs:
\begin{algorithm}[H]
	\caption{Fair Adversarial Networks: removing bias from data}
	\label{euclid}
	\begin{algorithmic}[1]
		\While {stopping criteria is not reached}
			\State Update the Autoencoder weights $W_A$ using the loss function \Statex\hspace{60pt} $\mathcal{L}_A(x,r,W_A,W_R)$.
			\State Obtain reconstructed data $y = \rho(x,W_A)$
			\For {k steps} 
				\State Update the Racist Network weights $W_R$ using the loss function
				\Statex\hspace{60pt}$\mathcal{L}_R(y,r,W_R,W_A)$.
			\EndFor
		\EndWhile
		
		\Return{y} \Comment{debiased data}

	\end{algorithmic}
\end{algorithm}

Given the success of this training procedure, the end result should be a dataset that is as similar to the original as possible while it should be harder/impossible to detect the undesirable protected characteristic. However, GANs are notoriously hard to train \cite{kodali2017convergence}. 

\subsection{Convergence}
Adversarial training often leads to very unstable or even run-away behaviour \cite{kodali2017convergence}.
Here we demonstrate acceptable convergence of our algorithm on five real-world datasets. While the convergence may seem still fairly sub-optimal, we argue that it is sufficiently good for our purpose. Crucially, we implement a ratchet mechanism which always preserves the state of the network with the lowest $\mathcal{L}_A$, therefore run-away behaviour after a period of convergence is not particularly problematic. 

While, we optimize the loss functions $\mathcal{L}_A$ and $\mathcal{L}_R$, they are not of interest for our purposes. Instead what we truly aim to achieve is to bring the predictive performance of the racist network after full training  $\bar{D}(\hat{r},\bar{r})$ down to random. For this task we operationalized $\bar{D}(\hat{r},\bar{r})$ as the best performance of the Racist Network on the validation data from 3 random initializations, and subsequent 10,000 epochs of training. The Racist Network used had a single hidden layer of the same width as the dataset.  The random benchmark we are using is $\bar{D}(ML(\bar{r}),\bar{r})$ i.e. always picking the most likely category. The other value of importance to us is  $MSE(x,y)$ as we aim to scramble the data as little as possible. Therefore these will be the focus of the analysis of convergence.

Further interesting values to observe include $D(\hat{r},\bar{r})$, the loss function of the racist network, but this does not provide a good indication of $\bar{D}(\hat{r},\bar{r})$, and therefore is fundamentally unsuitable to be a part of $\mathcal{L}_A$.  $\hat{D}(\hat{r},\bar{r})$ is also of interest which while being noisy provides good gradients for training as a part of $\mathcal{L}_A$.

\begin{figure}[H]
	\centering
	\includegraphics[width=9cm]{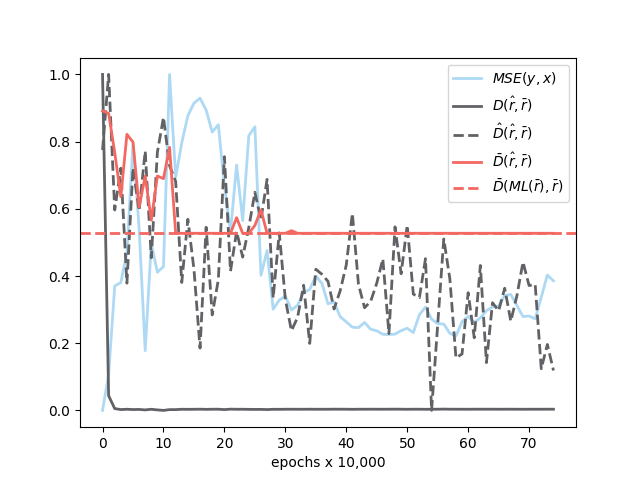}
	\caption{The Absenteeism at Work data includes personal information and total absenteeism time over 3 years for employees at a courier company in Brazil. We have selected age as the undesirable protected characteristic and successfully removed it. Source: UCI Machine Learning Database. Creators: Andrea Martiniano, Ricardo Pinto Ferreira, and Renato Jose Sassi. $\bar{D}$ values denote proportion of correct predictions of race, while all other values are arbitrarily scaled.}
	\label{fig:absenteeism}
\end{figure}

\begin{figure}[H]
	\centering
	\includegraphics[width=9cm]{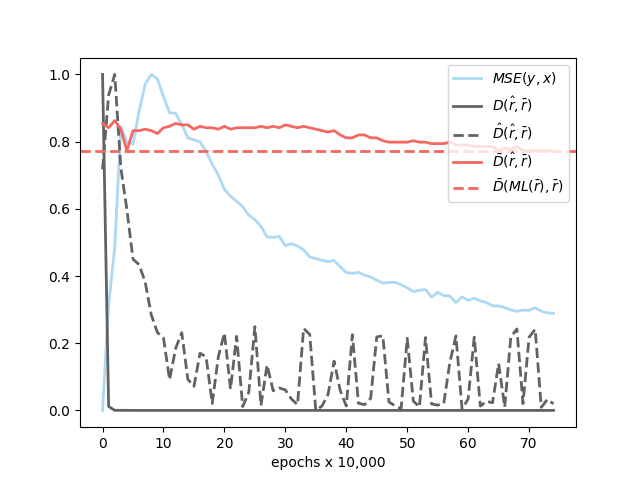}
	\caption{Performance data from schools in New York. The bias removed was a variable called 'Majority Black/Hispanic'. Source: Kaggle. Creators: PASSNYC. $\bar{D}$ values denote proportion of correct predictions of race, while all other values are arbitrarily scaled.}
	\label{fig:passnyc}
\end{figure}

\begin{figure}[H]
	\centering
	\includegraphics[width=9cm]{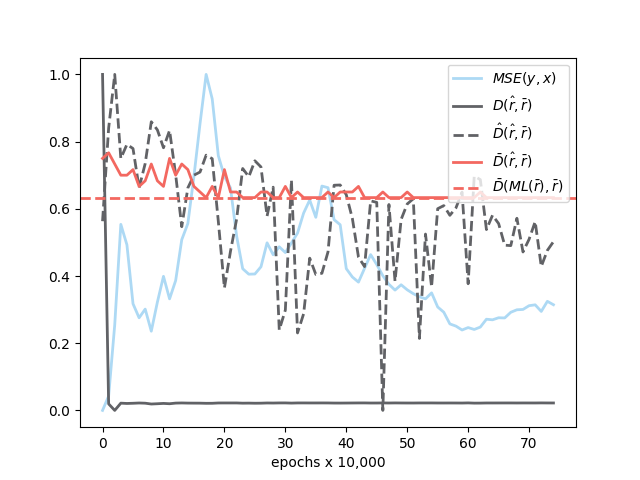}
	\caption{ The Heart Disease Dataset consists of blood measurements from a set of patients with and without heart disease. The bias variable removed was 'Sex'. Source: UCI Machine Learning Database. Creators: Hungarian Institute of Cardiology, University Hospital Zurich, University Hospital Basel, V.A. Medical Center Long Beach and Cleveland Clinic Foundation. $\bar{D}$ values denote proportion of correct predictions of gender, while all other values are arbitrarily scaled.}
	\label{fig:heart_disease}
\end{figure}

\begin{figure}[H]
	\centering
	\includegraphics[width=9cm]{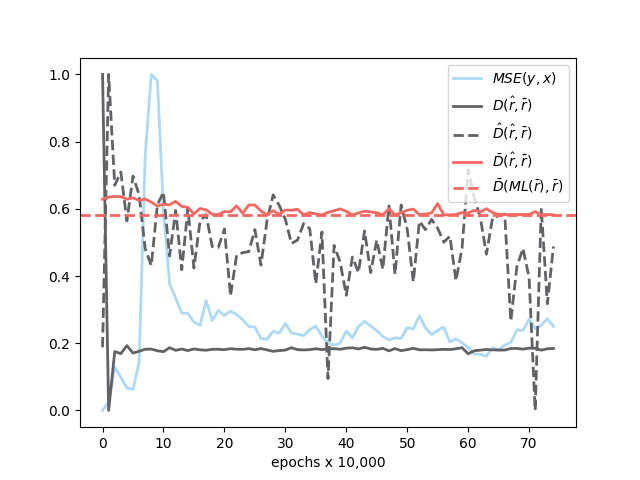}
	\caption{The COMPAS Dataset consists of profiles of criminals from Broward County, Florida. The bias variable removed was 'Race'. Source: Kaggle. Creators: ProPublica. $\bar{D}$ values denote proportion of correct predictions of race, while all other values are arbitrarily scaled.}
	\label{fig:compas}
\end{figure}

\begin{figure}[H]
	\centering
	\includegraphics[width=9cm]{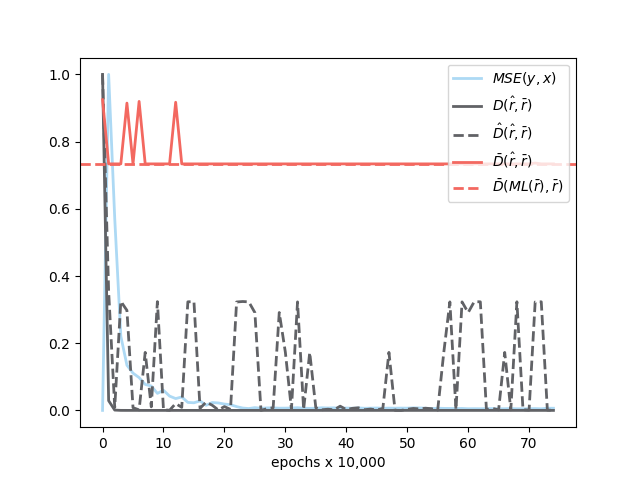}
	\caption{The Communities and Crime Dataset includes statistics on various communities i $\bar{D}$ values denote proportion of correct predictions of race, while all other values are arbitrarily scaled.n the U.S. Bias variable removed was 'Majority Black', indicating whether the community has a majority black population. Source: UCI Machine Learning Database. Creators: Michael Redmond.}
	\label{fig:communities}
\end{figure}

Figures \ref{fig:absenteeism} to \ref{fig:communities} clearly show that the $\bar{D}(\hat{r},\bar{r})$ has decreased to random performance and also an orderly behaviour of $MSE(x,y)$ correctly approaching minimum distortion of the data. The actual impact on Data Analytical outcomes will be discussed in another paper that is currently in preparation.

\newpage

\section{Discussion}
It is apparent that the outcomes of Data Analytics [DA] and Machine Learning [ML] are often perpetuating the human bias present in the datasets and therefore enacting illegal discrimination. Constraining the DA/ML outcomes to be fair is problematic as there is no universally accepted definition of fairness while at the same time many of the notions of fairness are very hard to implement, disrupting DA pipelines and putting a significant extra load on DA resources. One apparent solution is to remove bias from the data before proceeding with DA as usual; however, these methods generally, cannot account for non-linear, non-binary and/or multivariate relationships between race (or other biasing factor) and the rest of the data \cite{bolukbasi2016man}. This paper introduced \emph{Fair Adversarial Networks} [FANs] as a method that compensates for these shortcomings and provides a very significant improvement in both fairness and ease of use.

There is no universally accepted, or even legally binding, notion of fairness that can be used for optimisation, while at the same time many definitions of fairness are mutually exclusive. Unless a definition of fairness is provided by regulatory bodies it seems unlikely optimising for parity between groups on a fairness measure can be a useful bias-removal approach. Even if a mathematical notion of fairness becomes agreed
there is no guarantee that optimizing for parity on a training set achieves parity on the population-level.

Furthermore, the need to optimise for fairness introduces an extra term into the cost function of any optimisation procedure, which is not compatible with current DA tools. Data analysts are currently not required to write their own loss functions or optimisation procedures, therefore including such a requirement would damage their ability to perform their jobs. Even if the data analysts become comfortable with this requirement, the huge time overhead of this task makes it unlikely it will be performed in practise.  

Removing information about protected characteristics from the data is an attractive alternative. It's philosophically very simple - without knowledge of membership to a protected group (such as race) it should be impossible to discriminate based on it - therefore it removes the subjective nature of treating bias. It can also be made very simple, a single preprocessing step can remove bias from the data while the rest of DA/ML pipeline can remain exactly the same.

However many methods removing bias from data fail. Removing the column containing the protected characteristic is clearly insufficient due to the presence of proxy variables. A number of methods go beyond removing the column containing the protected characteristic and attempt to de-correlate the other characteristics from the protected ones. However, these approaches generally  cannot account for non-linear, non-binary, and/or multivariate relationships between the characteristics \cite{bolukbasi2016man}. To counter these problems this paper has introduced FANs. 

FANs are a version of adversarial networks with two main components 1) an \emph{autoencoder} that encodes a fair representation of data, and a \emph{Racist Network} which is the adversary predicting the protected characteristics (e.g. race) from the data, the performance of which needs to be minimised. 

The autoencoder's cost function  consist of reconstruction error of the transformed data, and also the performance of the racist network, both of which are to be minimised. The Racist Network simply tries to achieve the best predictive performance on the protected characteristic, using the autoencoder's output as its input. This system produces a data representation that is most similar to the original data, but at the same time from which the protected characteristics cannot (or at least are harder) be predicted. Any analytical methods can be subsequently used with such a representation. 

This paper describes the principles of FANs and demonstrates on five real-world, disparate datasets that FANs can indeed achieve their goal of removing the ability to predict the protected characteristic from the data, while minimising the difference between the original data and its fair reconstruction.

The problematic part of training FANs is that we aim to remove \emph{possibility to predict} protected characteristic from reconstructed data. Possibility to predict implies full training, not just the current state of the adversarial process. The success of our algorithm across five disparate datasets crucially relies on our approximation of full-training performance of a neural network from a single forward pass. While this approximation will remain our trade secret, it is possible to replicate our success on a one-off basis using conventional approaches and heavy parameter-tuning. 

This paper has demonstrated that FANs can consistently succeed at removing bias from datasets while keeping the necessary alterations of the data to the minimum. FANs are particularly valuable because they can be used as a generic and easy to use data pre-processing step, allowing all Data Analysts to account for biases in their datasets without significant overheads. 

\subsection*{Limitations and Future Directions}
%LIMITATIONS OF FANs

%no guarantee it won't make things worse
It is necessary to mention that under special circumstances FANs have the potential to make things worse for discriminated groups. FANs will remove all kinds of discrimination including the positive one, which might be a desirable way of breaking-up vicious cycles of deprivation in some areas.

%only compromise, not complete fairness
Optimising for two metrics at the same time is a compromise. The present paper has not attempted to analyse the residual discrimination which, while statistically insignificant, is likely still present. On the other hand the statistical insignificance can be seen as the criterion for success. Either way it is apparent that FANs provide a step in the right direction in respect to increasing the fairness.

% early-stopping & hyperparameters
Lastly, it is unclear what the correct time to stop the training of Neural Networks is, and what the right hyper-parameters are. It is certain that our neural architecture, neither hyper-parameter choice is optimal. Especially with GANs, one wishes to have interactive, optimal control of hyper-parameters throughout training to stabilize the process and ensure convergence. Therefore we are now exploring Reinforcement Learning as a method to control these factors interactively throughout training.

\bibliographystyle{plain}
\bibliography{lit}{}

\end{document}